\documentclass[runningheads]{llncs}
\usepackage{graphicx}
\usepackage{amssymb}
\usepackage{amsmath}
\usepackage{booktabs}
\usepackage{array}
\usepackage{float}
\usepackage{multirow}
\usepackage{siunitx}
\usepackage{stfloats}
\usepackage{pifont}
\usepackage[T1]{fontenc}
\usepackage{url}
\newcommand{\cmark}{\ding{51}}
\newcommand{\xmark}{\ding{55}}

\newcommand{\equalcontrib}{\textsuperscript{\textdagger}}
\newcommand{\correspondingauthor}{\textsuperscript{*}}
\newcolumntype{A}{S[table-format=2.1]}
\newcolumntype{E}{S[table-format=1.3]}
\newcolumntype{R}{S[table-format=2.1]}
\newcommand{\maketitlesupplementary}{%
  \begin{center}
    {\LARGE\bfseries Supplementary Material\par}
    \vspace{0.55em}
    {\large\bfseries Point-Selection Fine-Tuning Framework for Robust Point Cloud Classification\par}
    \vspace{0.65em}
    {\normalsize Da Li, Chang Ma, and Dongfu Yin\par}
  \end{center}
}
\begin{document}
\title{Point-Selection Fine-Tuning Framework for Robust Point Cloud Classification}
\titlerunning{Point-Selection Fine-Tuning for Robust Point Cloud Classification}
\author{Da Li\inst{1,2}\equalcontrib{} \and
Chang Ma\inst{2}\equalcontrib{} \and
Dongfu Yin\inst{1}\correspondingauthor{}\\[3pt]
{\small\mbox{\equalcontrib{}Equal contribution.\quad
\correspondingauthor{}Corresponding author.}}}
\authorrunning{D. Li et al.}
\institute{Guangdong Laboratory of Artificial Intelligence and Digital Economy
(Shenzhen), Shenzhen, China
\and
Shenzhen University, Shenzhen, China\\[3pt]
\email{li944104439@gmail.com; 2510235035@mails.szu.edu.cn}}
\maketitle

\begin{abstract}
Noisy and corrupted points can substantially degrade point cloud recognition performance, especially under challenging corruption settings.
In particular, full fine-tuning of 3D pre-trained models may amplify the influence of outliers and overwrite robustness priors learned during pre-training, while naive parameter-efficient adaptation remains sensitive to corrupted tokens.
To address this issue, we propose PSFT, a point-selection fine-tuning framework that improves robustness while remaining parameter-efficient.
PSFT first estimates point-wise influence from pre-pooling features and adaptively retains minimally influential points to suppress outliers.
Based on the selected subset, a prompt generation branch predicts layer-wise prompt tokens and injects them into a frozen backbone for lightweight downstream adaptation.
To further mitigate residual noise after selection, we append a lightweight feature filter with bottleneck MLP transformation and Beta-gated residual blending to refine patch-token representations before prediction.
Extensive experiments show that PSFT consistently reduces corruption error on ModelNet-C and ModelNet40-C across all tested 3D pre-trained backbones, while achieving the strongest ScanObjectNN-C results with ULIP-2 and Uni3D-B among the evaluated tuning strategies. Our implementation is available on: \url{https://github.com/CVChMA/PSFT/tree/master}.
\end{abstract}

\section{Introduction}
\label{sec:intro}

\begin{figure*}[!t]
    \centering
    \includegraphics[width=0.92\textwidth]{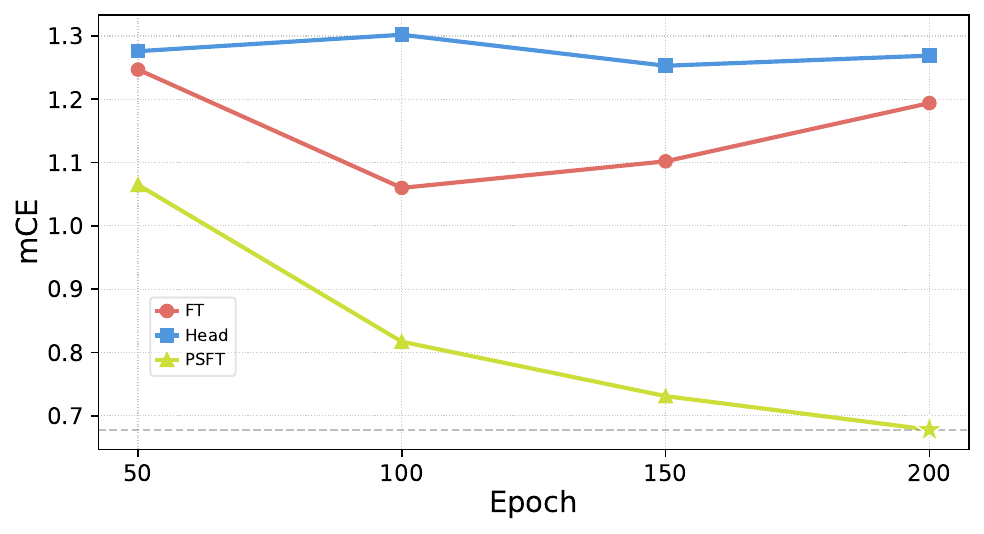}
    \caption{\textbf{Effect of Fine-Tuning Strategy on Robustness.} mCE denotes mean corruption error. FT updates all parameters, Head tunes only the classifier, and PSFT is our method.}
    \label{fig:intro}
\end{figure*}

Point cloud understanding is fundamental to applications ranging from 3D object recognition to autonomous-driving perception, where LiDAR--camera fusion supports accurate 3D detection~\cite{dstr}. However, noisy and corrupted points can severely degrade model performance, especially under challenging corruption settings such as ModelNet40-C~\cite{modelnet40-c}. Recent studies have made substantial progress on robust point cloud modeling~\cite{lgr-net,modelnet-c,pointcleannet,yan2020pointasnl}, but robust adaptation of 3D pre-trained models under corruption remains insufficiently explored. Our preliminary study on Point-MAE shows that full fine-tuning yields substantially higher corruption error than PSFT.

To analyze this behavior, we fine-tune pre-trained Point-MAE for 200 epochs and evaluate robustness every 50 epochs. We compare three strategies under the same schedule: full fine-tuning, head-only tuning, and PSFT. As shown in Figure~\ref{fig:intro}, full fine-tuning improves robustness at early epochs but later degrades, while head-only tuning remains stable yet with relatively high mCE. In contrast, PSFT consistently maintains the lowest corruption error. These results suggest that full fine-tuning can overwrite robust pre-trained priors, whereas head-only tuning lacks sufficient adaptation capacity.

We further evaluate IDPT~\cite{idpt}, a representative parameter-efficient fine-tuning (PEFT) baseline, on pre-trained Point-MAE and find that PEFT alone remains sensitive to corrupted inputs. This indicates that noisy points can still interfere with lightweight adaptation modules. Motivated by this observation, we adopt the point selection strategy of~\cite{refocusing} to filter input points before feeding them into both the PEFT branch and the frozen pre-trained network. Since residual noise may still remain after selection, we introduce a Feature Filter Module (FFM) that refines token features with a lightweight MLP and a Beta-gated residual connection, balancing denoising and information preservation.

The novelty of PSFT lies in this robustness-oriented coupling: selection protects both prompt generation and frozen-backbone inference, while feature filtering addresses noise that survives input selection.

We validate our method on three corruption benchmarks: ScanObjectNN-C~\cite{adaptpoints}, ModelNet-C~\cite{modelnet-c}, and ModelNet40-C~\cite{modelnet40-c}. 
Our objective is to improve corruption robustness while maintaining adaptation efficiency. 
To this end, we design lightweight modules that preserve robust priors in frozen pre-trained backbones. 
In addition, we provide targeted analyses to disentangle module-wise contributions and assess the transferability of robustness gains across benchmarks.
The primary contributions of our work can be summarized as follows:

\begin{itemize}
    \item We propose PSFT, a point-selection fine-tuning framework for robust 3D adaptation that combines point selection, prompt tuning on a frozen backbone, and feature filtering.

    \item PSFT achieves consistent robustness gains on ModelNet-C and ModelNet40-C across diverse 3D pre-trained backbones, and improves strong-backbone results on ScanObjectNN-C.

    \item We conduct systematic ablation and sensitivity analyses to quantify module-wise contributions and the impact of residual blending design choices on robustness.
\end{itemize}

\section{Related Work}

\label{sec:related}
\subsection{Point Cloud Classification}
Point cloud classification methods can be grouped into voxel-based, multi-view, and point-based paradigms. Voxel-based approaches discretize geometry into regular grids~\cite{voxel-base1}, while multi-view methods project 3D shapes into 2D renderings for image-style recognition~\cite{SimpleView}. Point-based methods process unordered points directly and avoid discretization artifacts, making them the dominant direction for fine-grained 3D recognition.

Within point-based modeling, the architecture trajectory has moved from shared-MLP designs to stronger local aggregation. PointNet~\cite{pointnet} establishes permutation-invariant feature extraction, PointNet++~\cite{pointnet++} introduces hierarchical neighborhoods, and DGCNN~\cite{dgcnn} improves relational modeling with dynamic graphs. Subsequent models such as PCT~\cite{pct}, PointMLP~\cite{pointmlp}, and PointNeXt~\cite{pointnext} further improve accuracy-efficiency trade-offs on standard classification benchmarks.

Recent progress is increasingly driven by pre-training and multimodal alignment. Transformer backbones improve long-range interaction modeling~\cite{pointtransformer,pointbert}. Masked or reconstruction-based pre-training further improves transferability~\cite{pointmae,2023recon}. Large-scale multimodal frameworks also strengthen 3D representation quality~\cite{pointgpt,ulip,ulip2,uni3d}. As backbones become stronger, adaptation under corrupted inputs becomes a central practical challenge.

\subsection{Parameter-Efficient Fine-Tuning in 3D Domain}
Parameter-efficient fine-tuning (PEFT) updates a small subset of parameters while keeping most backbone weights frozen. In the vision domain, representative paradigms include prompt tuning and adapter tuning, as exemplified by VPT~\cite{vpt} and AdaptFormer~\cite{visualadaptformer}. Their main advantage is reduced optimization cost with competitive transfer performance.

In 3D point cloud learning, PEFT has developed rapidly in recent years. IDPT~\cite{idpt} introduces instance-aware prompts for point cloud adaptation, Point-PEFT~\cite{pointpeft} combines point-prior prompts with geometry-aware adaptation, and PPT~\cite{ppt2024} explores efficient prompt-adapter integration. DAPT~\cite{dapt} provides a unified prompt-adapter design. More recent methods, such as GAPrompt~\cite{gaprompt} and vision-semantic prompting~\cite{visionsemanticprompt}, further incorporate geometric priors or external semantic guidance into frozen 3D backbones.

Despite strong clean-data transfer, most PEFT studies primarily optimize accuracy-efficiency trade-offs and report limited robustness analysis under corruption. Prior work also indicates that fine-tuning choices can induce nontrivial robustness--accuracy trade-offs~\cite{robustness_tradeoff_ft}. Motivated by this gap, our work focuses on robust PEFT for corrupted point clouds, rather than clean-data PEFT alone.

\subsection{Advances in 3D Point Cloud Model Robustness}
Robust point cloud research includes architecture-level invariance, denoising and outlier suppression, and certified defense. Geometry-aware methods improve invariance to structural perturbations~\cite{lgr-net}. Denoising-oriented approaches explicitly mitigate noisy observations~\cite{pointcleannet,yan2020pointasnl}. In adversarial settings, PointGuard studies certified robustness guarantees for 3D classification~\cite{pointguard}.

Unlike coordinate-space denoisers that reconstruct or resample point sets, PSFT couples influence-based selection with parameter-efficient adaptation and token-level refinement; the two directions are therefore complementary rather than interchangeable.

Another line investigates robustness from the representation perspective. Refocusing~\cite{refocusing} shows that highly influential points under pooling are often corrupted outliers, and suppressing their dominance can improve robustness. Robustness under incomplete or partial observations is also studied by transformer-based inference on partial inputs~\cite{transformer}, voting-based aggregation~\cite{voting}, and ensemble partial-point reasoning such as EPiC~\cite{levi2023epic}.

Robustness benchmarks provide standardized evaluation under diverse corruption settings. ModelNet-C~\cite{modelnet-c} and ModelNet40-C~\cite{modelnet40-c} emphasize controlled synthetic corruptions, while ScanObjectNN-C~\cite{adaptpoints} focuses on more realistic corruption patterns. These benchmarks consistently show that high clean-data transfer performance does not necessarily imply strong corruption robustness after adaptation. Our method is designed in this benchmark-driven setting, with emphasis on robust and parameter-efficient adaptation of frozen pre-trained models.

\section{Methodology}
\begin{figure*}[thbp]
    \centering
    \includegraphics[width=0.99\linewidth]{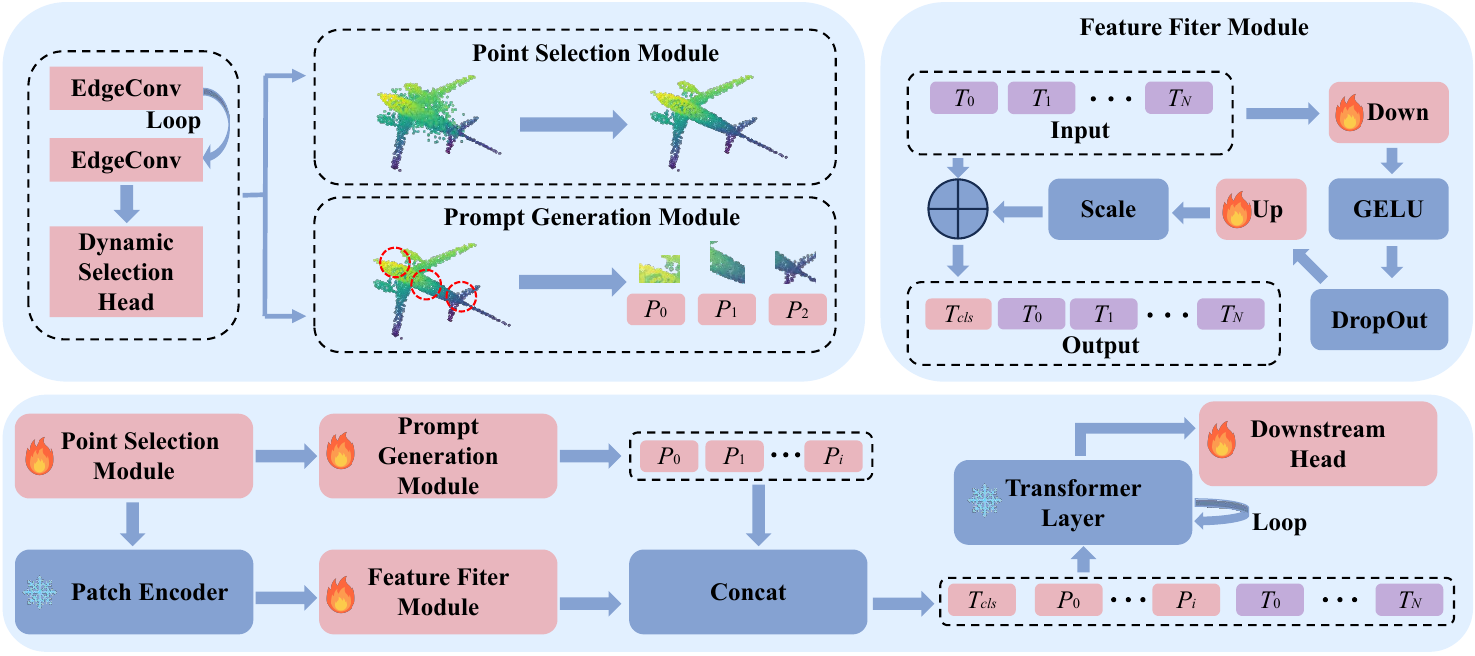}
    \caption{\textbf{Overview of PSFT.} The point selection module ranks points by contribution and retains a robust subset. The prompt generation module uses an EdgeConv-based branch to produce prompt tokens for a frozen transformer backbone. The feature filter module refines patch features by suppressing residual noise while preserving informative signals through residual blending.
    }
    \label{fig:construction}
   
\end{figure*}

\begin{figure}[t]
    \centering
    \includegraphics[width=0.92\linewidth]{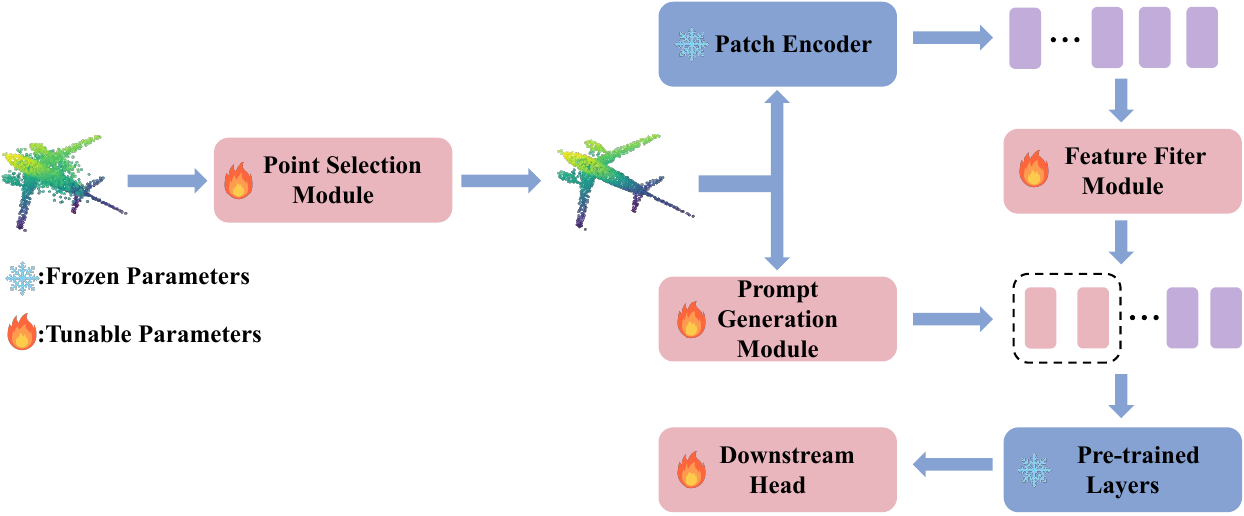}
    \caption{\textbf{PSFT Workflow.} PSFT first selects minimally influential points to suppress outliers, then generates layer-wise prompt tokens from the selected subset, and finally refines encoder features with a lightweight filter module before prediction.}
    \label{fig:workflow}
\end{figure}

We observe that full fine-tuning often provides limited robustness gains for 3D pre-trained models, and may even deteriorate performance under corruption. This suggests that the challenge is not only how to adapt the model to the downstream task, but also how to prevent corrupted points from destabilizing the adaptation process itself. Prior work has shown that outliers in corrupted point clouds can become highly influential under pooling and severely distort the resulting representation~\cite{refocusing}. Motivated by this, we propose a point-selection fine-tuning framework, PSFT, which explicitly reduces the impact of outliers while preserving the robust representations of the frozen backbone. Figure~\ref{fig:construction} and Figure~\ref{fig:workflow} illustrate the overall design.

Specifically, PSFT first applies a point selection module to retain minimally influential points and suppress outliers (Sec.~3.1). It then keeps the backbone frozen and introduces a prompt generation module for lightweight task adaptation while preserving the robustness of the pre-trained model (Sec.~3.2). Finally, a feature filter module refines token features after the frozen encoder before prediction (Sec.~3.3).

\subsection{Point Selection Module}
We adopt the refocusing strategy of~\cite{refocusing} as our point selection module. It serves two purposes: it ranks points according to their contribution to the pooled representation, and it adaptively determines how many points should be retained for subsequent processing.

\paragraph{Measuring Point-Wise Contribution via Max-Pooling.}
Let $X_f \in \mathbb{R}^{N \times \kappa}$ denote the final point-wise (or patch-wise) feature map before global pooling, where $N$ is the number of input elements and $\kappa$ is the channel dimension. For point-based backbones, $X_f$ is extracted from the last point feature map before pooling; for transformer backbones, it is taken from the patch-token feature map before the final pooling. We quantify the contribution of each point by counting how often it provides the maximum activation across feature channels.

Formally, the contribution score $I_F(j)$ for point $j \in \{1, \dots, N\}$ is defined as:

\begin{equation}
I_F(j) = \sum_{k=1}^\kappa \mathbb{I}\left(j = \mathop{\mathrm{arg\,max}}_{n \in \{1,\dots,N\}} X_f(n, k)\right),
\label{eq:point-contribution}
\end{equation}

where $\mathbb{I}(\cdot)$ is the indicator function. For each feature channel $k$, the point providing the maximum activation receives one vote. Hence, $I_F(j)$ reflects how strongly point $j$ influences the final pooled representation.

\paragraph{Adaptive Point Selection via Entropy-Guided Focus Estimation.}
To determine the appropriate number of points to retain in a corrupted point cloud, the entropy of point-wise feature importance is computed and used to guide the adaptive point selection process.

Following the previously defined point contribution score \( I_F(j) \), we obtain a contribution vector \( I_F = (I_F(1), \dots, I_F(N)) \). We normalize it into a probability distribution \( p \) over the point cloud:
\begin{equation}
    p_j = \frac{I_F(j)}{\sum_{i=1}^{N} I_F(i) + \varepsilon},
\end{equation}
where $\varepsilon$ is a small constant for numerical stability. We define $0 \log 0 = 0$ when computing entropy.

We then compute the entropy of this distribution:

\begin{equation}
    H(p) = -\sum_{j=1}^N p_j \ln(p_j),
\end{equation}

and normalize it to eliminate the influence of varying point cloud sizes:

\begin{equation}
    H_n(p) = \frac{H(p)}{\ln(N)}.
\end{equation}

The normalized entropy \( H_n(p) \in [0, 1] \) measures how uniformly the network distributes its focus across the point cloud. A higher entropy indicates that the representation depends on many points, while a lower entropy suggests that only a small subset dominates the pooled feature.

To adaptively determine how many points should be retained, we define the selected point number \( S \) as:

\begin{equation}
    S = \mathrm{clip}\!\left(\mathrm{round}\!\left((1 - H_n(p)) \cdot N\right), 1, N\right).
\end{equation}

After computing \(S\), we sort points in ascending order of \(I_F(j)\) and retain the first \(S\) minimally influential points. This strategy follows the observation of~\cite{refocusing} that highly influential points are more likely to correspond to corrupted outliers. When the representation is highly concentrated (low entropy), we preserve more points to avoid over-trusting a very small subset; when the representation is more dispersed, a smaller robust subset is typically sufficient.

Ignoring rounding and clipping, the retained ratio is $S/N=1-H_n(p)$; hence a change $\Delta H_n$ changes the ratio by $-\Delta H_n$. This bounded monotone rule makes the selection strength adaptive rather than dataset-fixed.

\subsection{Prompt Generation Module}

To adapt the model while preserving the robustness of the pre-trained backbone, we adopt a prompt-based tuning strategy that inserts one learnable prompt token into each transformer layer. The prompts are conditioned on the selected point subset through an EdgeConv-based prompt generator~\cite{dgcnn}, which captures local geometric structure and produces a compact global descriptor for prompt prediction.

Unlike full fine-tuning, PSFT keeps the backbone frozen rather than updating its parameters, since modifying the pre-trained model may compromise the robustness it has already learned. Instead, PSFT performs task adaptation through a prompt branch that predicts layer-specific prompt vectors from the selected points, while the frozen encoder branch continues to produce the patch tokens. This design introduces task-specific flexibility while minimizing interference with the robust representations of the backbone.

Formally, let \( T_{\text{cls}} \in \mathbb{R}^{1 \times d} \) denote the classification token, let \(T_i \in \mathbb{R}^{M \times d}\) denote the patch tokens entering the $i$-th transformer layer, and let \(P_i \in \mathbb{R}^{1 \times d}\) denote the prompt token predicted for that layer. The prompt-augmented token sequence processed by the $i$-th transformer layer \(L_i\) is

\begin{equation}
    x_i = L_i\left([T_{\text{cls}}, P_i, T_i]\right).
\end{equation}

This formulation inserts exactly one prompt token per layer and avoids cumulative prompt growth. By conditioning \(P_i\) on the selected points, PSFT adapts the model to the downstream task without modifying the frozen backbone, thereby better preserving robustness under corruption.

\subsection{Feature Filter Module}

Although the Point Selection Module filters out a significant portion of corrupted or less informative points, the retained subset may still contain residual noise due to sampling imperfections or perturbations such as jitter. To further improve representation quality, we introduce a Feature Filter Module (FFM) after the frozen encoder to refine patch-token features before prediction.

The Feature Filter Module is a lightweight funnel-shaped multi-layer perceptron (MLP) applied to patch tokens after the frozen encoder and before the final pooling stage. We keep the classification token unchanged and apply the filter token-wise to each patch token. For an input token \( x_o \in \mathbb{R}^d \), the filtered token \( x_f \in \mathbb{R}^d \) is obtained as

\begin{equation}
x_f = \text{DropOut}(\text{GELU}(x_o \cdot W_{\text{down}})) \cdot W_{\text{up}},
\end{equation}

where \( W_{\text{down}} \in \mathbb{R}^{d \times d'} \) and \( W_{\text{up}} \in \mathbb{R}^{d' \times d} \) are projection matrices with a bottleneck dimension \( d' < d \), and GELU and Dropout are used to introduce non-linearity and regularization.

While filtering suppresses noise, it may also weaken useful cues. To balance this trade-off, we blend the filtered and original features through a residual connection. During training, the blending ratio is sampled as \( \alpha \sim \mathrm{Beta}(\lambda,\lambda) \); one \(\alpha\) is sampled per input sample and shared across its tokens.

\begin{equation}
x_{\text{final}} = (1 - \alpha) \cdot x_f + \alpha \cdot x_o.
\end{equation}

The Beta distribution serves two purposes. First, it provides a bounded interpolation mechanism in \([0,1]\), allowing smooth control over the contribution of the filtered and original features. Second, the stochasticity introduced during training acts as an implicit regularizer, improving robustness to corrupted inputs. In the analysis of Sec.~4, we examine how different training-time \(\lambda\) settings and test-time \(\alpha\) values affect this trade-off.

For $\alpha\sim\mathrm{Beta}(\lambda,\lambda)$, $\mathbb{E}[\alpha]=1/2$ and $\mathrm{Var}(\alpha)=1/[4(2\lambda+1)]$, so smaller $\lambda$ exposes training to a broader range of mixtures. Moreover, $x_{\mathrm{final}}-x_o=(1-\alpha)(x_f-x_o)$, explicitly bounding the correction applied to the frozen feature.

\section{Experiments}
\subsection{Datasets and Basic Settings}
\paragraph{Datasets.}
We evaluate our method on three representative corruption benchmarks: ModelNet-C, ModelNet40-C, and ScanObjectNN-C. These datasets cover both synthetic and real-world corruption settings, enabling a comprehensive assessment of robustness under different degradation conditions.

\paragraph{Implementation details.}
Models are trained on the clean versions of the corresponding datasets, such as ModelNet40~\cite{modelnet40} and ScanObjectNN~\cite{Scanobjectnn}. Evaluations are then conducted on their corruption test suites, namely ModelNet-C~\cite{modelnet-c}, ModelNet40-C~\cite{modelnet40-c}, and ScanObjectNN-C~\cite{adaptpoints}. We use the AdamW optimizer with a learning rate of $5\times10^{-4}$ and train for 300 epochs. For ModelNet-C and ScanObjectNN-C, we follow the original benchmark protocols and report mean Corruption Error (mCE, $\downarrow$). For ModelNet40-C, we report mean Error Rate (mER, $\%,\downarrow$), i.e., the average classification error over all corruption types in the benchmark. Unless otherwise specified, the main paper reports results without data augmentation so that the effect of robust adaptation can be isolated more clearly; results with WOLFMix~\cite{modelnet-c} augmentation are deferred to the supplementary material.

\subsection{Comparison Results}

\paragraph{Overall Performance.}
We evaluate PSFT on three corrupted point cloud benchmarks: ModelNet-C, ModelNet40-C, and ScanObjectNN-C. As shown in Table~\ref{tab:modelnet-c}, Table~\ref{tab:modelnet40-c}, and Table~\ref{tab:scanobjectnn-c}, PSFT improves robustness on all four evaluated backbones on ModelNet-C and ModelNet40-C, and on the stronger ULIP-2 and Uni3D-B backbones on ScanObjectNN-C. The evaluated pre-trained backbones include PointBERT~\cite{pointbert}, PointMAE~\cite{pointmae}, Uni3D~\cite{uni3d}, and ULIP-2~\cite{ulip,ulip2}.

\paragraph{Results on ModelNet-C and ModelNet40-C}
Table~\ref{tab:modelnet-c} shows that PSFT consistently reduces mCE for all four evaluated pre-trained backbones on ModelNet-C. The best result among the compared methods is achieved by ULIP-2 with PSFT, which reaches an mCE of 0.530. Table~\ref{tab:modelnet40-c} shows a similar trend on the broader ModelNet40-C suite: PSFT substantially lowers mER for strong backbones such as ULIP-2 and Uni3D-B, indicating that the proposed tuning strategy generalizes across both synthetic corruption benchmarks. The corresponding results with augmentation are reported in the supplementary material.

\paragraph{Results on ScanObjectNN-C.}
On the more challenging real-world benchmark ScanObjectNN-C, PSFT achieves the best mCE among the compared PEFT variants, reaching 0.685 on ULIP-2. The improvements obtained with ULIP-2 and Uni3D-B support the robustness benefit of point selection and feature filtering on strong backbones, while the PointBERT and PointMAE results indicate that the gains do not transfer uniformly to every backbone in real-world settings.

Relative to backbone-specific full fine-tuning, the ScanObjectNN-C mCE changes are $+0.039$, $+0.119$, $-0.092$, and $-0.099$ for Point-BERT, Point-MAE, ULIP-2, and Uni3D-B, respectively. This mixed pattern contrasts with the uniform synthetic-benchmark gains and indicates backbone-dependent transfer under real corruptions.

\begin{table}[!t]
  \tiny
  \centering
  \caption{\textbf{Experimental Results on ModelNet-C.} OA and mCE denote overall accuracy and mean corruption error, respectively.} 
  \resizebox{\columnwidth}{!}{%
\begin{tabular}{l|c|c|c|c|c|c|c|c|c}
    \hline
    Model & OA ↑ & mCE ↓ & Scale & Jitter & Drop-G & Drop-L & Add-G & Add-L & Rotate   \\
    
    \hline
    PointNet~\cite{pointnet} & 90.7\% & 1.422 & 1.266 &  0.642 & 0.500 & 1.072 & 2.980 & 1.593 & 1.902\\
   
    RSCNN~\cite{RSCNN} & 92.3\% & 1.130 & 1.074 & 1.171 & 0.806 &  1.517 & 0.712 & 1.153 &  1.479\\
    
    PAConv~\cite{PAConv} & 93.6\% & 1.104 & 0.904 & 1.465 & 1.000 &  1.005 & 1.085 & 1.298 &  0.967\\
     PointNet++~\cite{pointnet++} & 93.0\% & 1.072 & 0.872 &  1.177 & 0.641 & 1.802 & 0.614 & 0.993 & 1.405\\
     SimpleView~\cite{SimpleView} & \textbf{93.9}\% & 1.047 & 0.872 & 0.715 & 1.242 &  1.357 & 0.983 & 0.844 &  1.316\\
    
    Point-DAE~\cite{PointDAE} & 93.3\% & 0.835 & 0.957 &  0.883 & 0.944 & 0.841 & 0.668 & 0.815 & 0.736 \\
APCT~\cite{APCT} & - & 0.722 & 0.947 & 0.883  & 0.468 & 0.850 & 0.285 & \textbf{0.298} & 1.326 \\
    \hline
    Point-BERT~\cite{pointbert}  & 92.2\% & 1.098 & 0.872 &  1.503 & 0.512 & 0.903 & 1.085 & 1.495 & 1.316 \\
    w. IDPT~\cite{idpt} & 91.3\% & 1.142 & 0.936 & 0.813 & 0.512 & 1.174 & 2.071 & 1.353 & 1.135\\
    w. PSFT (Ours) & 91.4\% & 0.790 & 1.149 & 0.503 & 0.383   & 0.961 & 0.319 & 0.625 & 1.591\\

    \hline
    Point-MAE~\cite{pointmae}  & 92.6\% & 0.983 & 0.840 & 1.256 & 0.464   & 0.763 & 1.075 & 1.265 & 1.219 \\
    w. IDPT~\cite{idpt} & 92.7\% & 1.008 & 0.851 & 0.769 & 0.613 & 1.000 & 1.346 & 1.145 & 1.330
\\
    w. PSFT (Ours) & 91.9\% & 0.704 & 1.032 & \textbf{0.427}  & 0.431 & 0.860 & 0.298 & 0.502 & 1.377\\

    \hline
    ULIP-2~\cite{ulip2}  & 92.1\% & 0.863 & 0.851 & 1.778 & 0.492 & \textbf{0.734} & 0.386 & 0.862 & 0.935 \\
     w. IDPT~\cite{idpt} & 92.9\% & 0.775 & \textbf{0.745} & 1.709 & 0.323 & 0.947 & 0.454 & 0.851 & 0.395\\
      w. PSFT (Ours) & 93.3\% & \textbf{0.530} & 0.819 & 0.585 & \textbf{0.298} & 0.816 & \textbf{0.244} & 0.567 & \textbf{0.381} \\

    \hline
    Uni3d-B~\cite{uni3d}  & 92.9\% & 0.746 & 0.777 & 1.228  & 0.415 & 0.787 & 0.444 & 0.833 & 0.740\\
     w. IDPT~\cite{idpt} & 91.5\% & 0.946 & 1.000 & 1.472 & 0.653 & 1.266 & 0.569 & 1.142 & 0.521\\
 w. PSFT (Ours) & 93.4\% & 0.566 & 0.862 & 0.459 & 0.327 & 0.986 & 0.254 & 0.575 & 0.498\\

    \hline
  \end{tabular}}
  
  \label{tab:modelnet-c}
  \vspace{2pt}
\end{table}

\begin{table*}[!t]
\small
\renewcommand{\arraystretch}{1.14}
\setlength{\tabcolsep}{3.5pt}
\caption{\textbf{Error Rates on ModelNet40-C.} mER denotes mean error rate.}
\label{tab:modelnet40-c}
\centering
\makebox[\textwidth][c]{%
\resizebox{1.03\textwidth}{!}{%
\begin{tabular}{l||c|ccccc|ccccc|ccccc}
\specialrule{1pt}{1.1pt}{1pt}
& & \multicolumn{5}{c|}{Density Corruptions} & \multicolumn{5}{c|}{Noise Corruptions} & \multicolumn{5}{c}{Transformation Corruptions}\\
\cline{3-17}
Model & mER $\downarrow$ & \shortstack{Occlu-\\sion} & LiDAR & \shortstack{Density\\Inc.} & \shortstack{Density\\Dec.} & Cutout & Uniform & Gaussian & Impulse & \shortstack{Up-\\sampling} & \shortstack{Back-\\ground} & Rotation & Shear & FFD & RBF & \shortstack{Inv.\\RBF}\\
\noalign{\global\arrayrulewidth1pt}\hline\noalign{\global\arrayrulewidth0.4pt}
PointNet~\cite{pointnet} & 28.3 & 52.3 & 54.9 & 10.5 & 11.6 & 12.0 & 12.4 & 14.4 & 29.1 & 14.0 & 93.6 & 36.8 & 25.4 & 21.3 & 18.6 & 17.8 \\
SimpleView~\cite{SimpleView} & 27.2 & 55.5 & 82.2 & 13.7 & 17.2 & 20.1 & 14.5 & 14.2 & 24.6 & 17.7 & 46.8 & 30.7 & 18.5 & 17.0 & 17.9 & 17.2 \\
RSCNN~\cite{RSCNN} & 26.2 & 51.8 & 68.4 & 16.8 & 13.2 & 13.8 & 24.6 & 18.3 & 46.2 & 20.1 & 18.3 & 29.2 & 17.0 & 18.1 & 19.2 & 18.6 \\
DGCNN~\cite{dgcnn} & 25.9 & 59.2 & 81.0 & 14.1 & 17.3 & 15.4 & 14.6 & 16.6 & 24.9 & 19.1 & 53.1 & 19.1 & 12.1 & 13.1 & 14.5 & 14.0 \\
PCT~\cite{pct} & 25.5 & 56.6 & 76.7 & 11.8 & 14.3 & 14.5 & 12.1 & 13.9 & 39.1 & 17.4 & 57.9 & 18.1 & 11.5 & 12.4 & 13.0 & 12.6 \\
PointNet++~\cite{pointnet++} & 23.6 & 54.7 & 66.5 & 16.0 & 10.0 & 10.7 & 20.4 & 16.4 & 35.1 & 17.2 & 18.6 & 27.6 & 13.4 & 15.2 & 16.4 & 15.4 \\
\hline
Point-BERT~\cite{pointbert} & 26.5 & 54.7 & 73.8 & 10.8 & 13.4 & 13.6 & 17.5 & 22.8 & 46.7 & 23.8 & 36.7 & 25.3 & 13.3 & 14.5 & 15.9 & 14.9\\
w. IDPT~\cite{idpt} & 26.7 & 59.3 & 80.8 & 10.7 & 14.1 & 13.8 & 11.9 & 13.7 & 32.7 & 14.6 & 72.4 & 21.8 & 12.9 & 14.3 & 14.1 & 12.8\\
w. PSFT (Ours) & 20.0 & 56.9 & 70.9 & 9.2 & 10.6 & 9.6 & 11.1 & 11.2 & 10.4 & 10.2 & 9.9 & 30.7 & 16.9 & 15.5 & 14.1 & 13.5\\
\hline
Point-MAE~\cite{pointmae} & 23.9 & 53.5 & 74.8 & 10.7 & 11.3 & 11.9 & 15.7 & 18.5 & 33.0 & 18.8 & 35.5 & 22.7 & 11.5 & 13.1 & 13.9 & 13.0\\
w. IDPT~\cite{idpt} & 24.6 & 55.3 & 79.5 & 11.1 & 13.3 & 13.8 & 11.0 & 12.9 & 18.0 & 13.4 & 57.2 & 26.4 & 13.5 & 15.0 & 14.7 & 13.8\\
w. PSFT (Ours) & 18.3 & 55.5 & 66.6 & 9.0 & 10.7 & 10.1 & 10.5 & \textbf{10.3} & 9.4 & 9.6 & 9.6 & 24.3 & 12.9 & 12.9 & \textbf{12.0} & \textbf{11.3}\\
\hline
ULIP-2~\cite{ulip2} & 24.1 & 51.4 & 80.0 & 9.7 & 10.8 & 10.9 & 29.9 & 35.7 & 18.7 & 31.9 & 12.9 & 18.6 & 10.5 & 12.1 & 14.7 & 13.8\\
w. IDPT~\cite{idpt} & 23.3 & \textbf{42.6} & 79.1 & 10.0 & 11.6 & 12.3 & 29.7 & 37.4 & 15.4 & 32.0 & 16.1 & 8.2 & 7.8 & 12.8 & 17.6 & 16.4\\
w. PSFT (Ours) & 14.9 & 44.6 & 53.3 & \textbf{7.5} & \textbf{8.8} & \textbf{9.0} & 12.0 & 12.8 & \textbf{8.2} & \textbf{9.5} & \textbf{7.6} & \textbf{7.6} & \textbf{7.1} & \textbf{10.4} & 12.5 & 11.9\\
\hline
Uni3d-B~\cite{uni3d} & 18.9 & 44.2 & 58.2 & 9.6 & 10.3 & 11.3 & 15.1 & 18.9 & 20.2 & 21.6 & 14.1 & 13.4 & 8.8 & 11.7 & 13.4 & 12.6\\
w. IDPT~\cite{idpt} & 23.0 & \textbf{38.2} & 56.3 & 10.5 & 11.8 & 16.7 & 22.0 & 29.3 & 26.4 & 33.2 & 17.3 & 10.0 & 10.7 & 18.1 & 22.8 & 22.1\\
w. PSFT (Ours) & \textbf{14.2} & 41.2 & \textbf{42.4} & 8.1 & 9.8 & 10.5 & \textbf{10.0} & 10.8 & 8.9 & 9.7 & 7.6 & 9.0 & 7.5 & 11.1 & 13.3 & 12.7\\
\noalign{\global\arrayrulewidth1pt}\hline\noalign{\global\arrayrulewidth0.4pt}
\end{tabular}%
}}
\vspace{2pt}
\end{table*}

\begin{table}[!t]
  \centering
  \caption{\textbf{Experimental Results on ScanObjectNN-C.} OA and mCE denote overall accuracy and mean corruption error, respectively.}
   \resizebox{\columnwidth}{!}{%
\begin{tabular}{l|c|c|c|c|c|c|c|c|c}
    \hline
    Model & OA ↑ & mCE ↓ & Scale & Jitter & Drop-G & Drop-L & Add-G & Add-L & Rotate   \\
    
    \hline
    
    RPC~\cite{modelnet-c} & 74.7\% & 1.326 & 1.317 & 1.073 & 1.455 &  1.305 & 1.142 & 1.587 &  1.402\\
    PointNet++~\cite{pointnet++} & 86.2\% & 0.969 & 0.897 &  1.103 & 0.550 & 1.277 & 0.947 & 0.905 & 1.107\\
    PointNeXt~\cite{pointnext} & 87.3\% & 0.921 & 0.803 & 1.079 & 0.807 & 0.942 & 0.944 & 0.875 &  0.995\\
    PointGPT~\cite{pointgpt} & - & 0.847 & 1.199 & 1.040 & 0.585 & 0.769 & 0.361 & 0.722 & 1.251 \\
    APCT~\cite{APCT} & - & 0.742 & 1.045 & 0.989  & 0.489 & 0.670 & \textbf{0.300} & \textbf{0.630} & 1.071 \\
    
    \hline
    Point-BERT~\cite{pointbert}  & 83.4\% & 0.931 & 0.618 & 1.088 & 0.587 & 0.736 & 0.661 & 1.630 & 1.195
\\
    w. IDPT~\cite{idpt} & 83.6\% & 1.053 & 0.749 & 0.958 & 0.545 & 0.851 & 1.117 & 2.154 & 0.996 \\
    w. PSFT (Ours) & 79.4\% & 0.970 & 0.898 & 0.954 & 0.608 & 0.937 & 0.493 & 1.612 & 1.288
\\

    \hline
    Point-MAE~\cite{pointmae}  & 84.3\% & 0.895 & 0.609 & 0.952 & 0.569 & 0.726 & 0.550 & 1.722 & 1.135
\\
    w. IDPT~\cite{idpt} & 85.9\% & 0.982 & 0.585 & 0.910 & 0.511 & 0.733 & 1.126 & 2.044 & 0.963\\
    w. PSFT (Ours) & 77.6\% & 1.014 & 0.960 & 0.974 & 0.701 & 0.990 & 0.580 & 1.568 & 1.326
\\

    \hline
    ULIP-2~\cite{ulip2}  & 86.2\% & 0.777 & 0.538 & 1.096 & 0.524 & \textbf{0.637} & 0.376 & 1.348 & 0.918
 \\
   w. IDPT~\cite{idpt} & \textbf{87.3}\% & 0.792 & 0.533 & 1.031 & \textbf{0.399} & 0.673 & 0.654 & 1.670 & \textbf{0.584}\\
      w. PSFT (Ours) & 86.1\% & \textbf{0.685} & 0.621 & 0.960 & 0.431 & 0.670 & 0.333 & 1.079 & 0.700
 \\

    \hline
    Uni3d-B~\cite{uni3d}  & 85.8\% & 0.801 & \textbf{0.509} & 1.085 & 0.532 & 0.680 & 0.365 & 1.502 & 0.933
\\
w. IDPT~\cite{idpt} & 83.6\% & 0.946 & 0.713 & 1.138 & 0.696 & 0.967 & 0.491 & 1.771 & 0.846\\
 w. PSFT (Ours) & 86.4\% & 0.702 & 0.626 & \textbf{0.871} & 0.458 & 0.733 & 0.341 & 1.079 & 0.805
\\

    \hline
  \end{tabular}}
  
  \label{tab:scanobjectnn-c}
\end{table}

\subsection{Further Analysis}

\paragraph{Impact of Each Component.}
We conduct a series of controlled experiments to investigate the impact of each component of PSFT, as summarized in Table~\ref{tab:abla}.
We observe that each module contributes to robustness improvements over the full fine-tuning baseline. In particular, using only the point selection module already yields a clear gain on all three benchmarks, and adding either prompt generation or feature filtering further improves robustness. Combining all three modules gives the strongest overall performance on ModelNet-C and ModelNet40-C, reaching an mCE of 0.530 and an mER of 14.9, while on ScanObjectNN-C the PS+FFM variant achieves the lowest mCE, indicating a slightly different trade-off on real-world corruptions.

\begin{figure}[!t]
    \centering
    \includegraphics[width=0.74\linewidth]{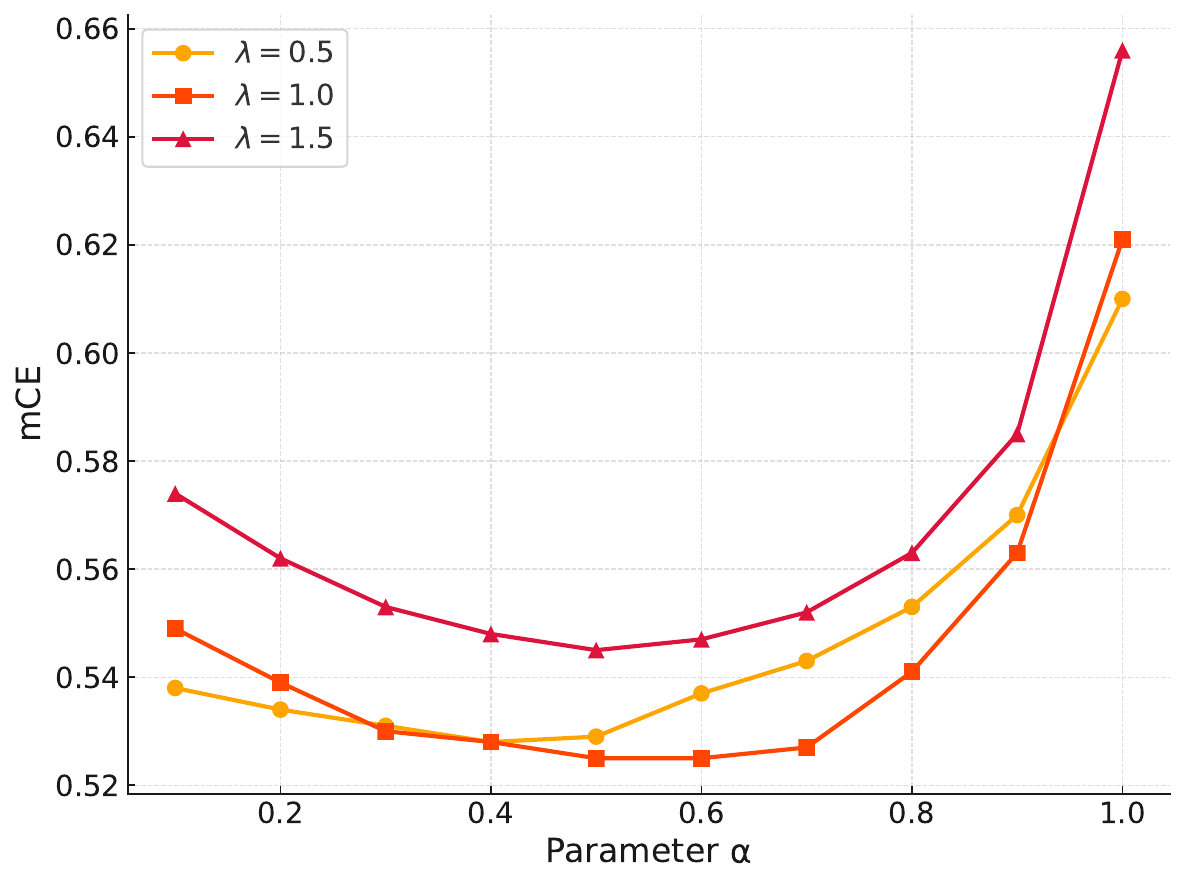}
    \caption{\textbf{Effect of Test-Time Blending Coefficient $\alpha$.} The plot shows mCE versus test-time $\alpha$ under training-time sampling distributions $\mathrm{Beta}(\lambda,\lambda)$ with $\lambda \in \{0.5, 1.0, 1.5\}$. Across all settings, robustness is best around $\alpha \approx 0.5$, and training with $\mathrm{Beta}(0.5,0.5)$ yields the lowest overall mCE.}
    \label{fig:effectofalpha}
\end{figure}

\paragraph{Effect of Test-Time $\alpha$ under Different Training-Time $\lambda$ Settings.}
Figure~\ref{fig:effectofalpha} illustrates how the test-time blending parameter $\alpha$ affects robustness (mCE) under three training-time Beta distributions: $\text{Beta}(0.5,0.5)$, $\text{Beta}(1.0,1.0)$, and $\text{Beta}(1.5,1.5)$. For clarity, the key observations are:

\noindent\textbf{(i) Optimal test-time $\alpha$ is near 0.5 across all settings}: mCE is minimized when $\alpha \approx 0.5$, indicating that a balanced residual mixture of the identity and transformed features ($x$ and $f(x)$) yields the strongest robustness.\par
\noindent\textbf{(ii) $\text{Beta}(0.5,0.5)$ gives the strongest overall robustness profile}: assigning more probability mass near boundaries exposes training to broader residual-mixing patterns and consistently lowers mCE.\par
\noindent\textbf{(iii) Robustness degrades as $\boldsymbol{\alpha \rightarrow 1.0}$}: across all training configurations, rapidly increasing mCE suggests that excessive reliance on the identity path weakens the corrective effect of $f(x)$ during inference.\par
\noindent\textbf{(iv) Larger $\lambda$ narrows residual-mixing diversity}: compared with $\lambda=0.5$, $\text{Beta}(1.0,1.0)$ and $\text{Beta}(1.5,1.5)$ correspond to weaker robustness, indicating that reducing extreme residual weights is less effective than broader mixing exposure.\par
Overall, stochastic residual blending with broad training-time coverage is beneficial, and test-time $\alpha \approx 0.5$ provides a stable trade-off between denoising and information preservation.

\paragraph{Comparison with Other Fine-Tuning Strategies.}
We conduct experiments on ModelNet-C~\cite{modelnet-c} using ULIP-2 to compare PSFT with representative PEFT baselines, including VPT~\cite{vpt} and AdaptFormer~\cite{visualadaptformer}. As shown in Table~\ref{tab:comparison_tuning}, AdaptFormer reduces the number of tunable parameters but still trails full fine-tuning in robustness, while VPT performs substantially worse under corruption. In contrast, PSFT improves robustness while preserving parameter efficiency. Compared to full fine-tuning, PSFT reduces the number of trainable parameters from 22.1M to 2.4M and lowers mCE from 0.863 to 0.530.

Thus, PSFT uses 10.9\% of the trainable parameters of full fine-tuning, corresponding to an 89.1\% reduction.

\begin{table}[!t]
\scriptsize
\setlength{\tabcolsep}{5mm}

\centering
\caption{\textbf{Comparison with Fine-Tuning Baselines on ULIP-2 (ModelNet-C).} \#TP denotes the number of trainable parameters.} 
\label{tab:comparison_tuning}
\setlength{\tabcolsep}{10pt}
\resizebox{0.65\columnwidth}{!}{ \begin{tabular}{ lcc }
\toprule
Method  & \#TP (M) & mCE \\
\midrule
w. FT  & 22.1  & 0.863 \\
w. VPT-Deep & 0.4 & 1.454 \\
w. AdaptFormer & 0.9 & 0.920 \\
w. \textbf{PSFT (ours)} & 2.4 & \textbf{0.530} \\
\bottomrule
\vspace{-8pt}
\end{tabular}
}
\end{table}

\begin{table}[!t]
  \centering
  \caption{\textbf{Ablation Study on ModelNet-C~\cite{modelnet-c}, ModelNet40-C~\cite{modelnet40-c}, and ScanObjectNN-C~\cite{Scanobjectnn}.} PS, PG, and FFM denote the point selection, prompt generation, and feature filter modules, respectively.}
  \setlength{\tabcolsep}{3pt} 
  \renewcommand{\arraystretch}{0.9}
  \scriptsize
  \begin{tabular}{ccc|cc|c|cc}
    \toprule
    \multicolumn{3}{c|}{Components} & \multicolumn{2}{c}{ModelNet-C~\cite{modelnet-c}} & \multicolumn{1}{c}{ModelNet40-C~\cite{modelnet40-c}} &
    \multicolumn{2}{c}{ScanObjectNN-C~\cite{Scanobjectnn}} \\
    \cmidrule(lr){1-3}\cmidrule(lr){4-5}\cmidrule(lr){6-6}\cmidrule(lr){7-8}
    PS & PG & FFM &
    OA ↑ & mCE ↓ & mER $\downarrow$ &
    OA ↑ & mCE ↓ \\
    \midrule
    \xmark & \xmark & \xmark & 92.1\% & 0.863 & 24.1 & 86.2\% & 0.777 \\
    \xmark & \xmark & \cmark & 93.6\% & 0.819 & 24.6 & 87.4\% & 0.735 \\

    \xmark & \cmark & \xmark & 92.6\% & 0.821 & 20.3 & 85.3\% & 0.817 \\
     \cmark & \xmark & \xmark & 91.9\% & 0.656 & 17.6 & 85.4\% & 0.724 \\
    \xmark & \cmark & \cmark & 93.4\% & 0.734 & 19.9 & 86.9\% & 0.758 \\

    \cmark & \xmark & \cmark & \textbf{94.1}\% & 0.558 & 15.8 & \textbf{87.4}\% & \textbf{0.654} \\

    \cmark & \cmark & \xmark & 92.4\% & 0.648 & 16.8 & 84.5\% & 0.780 \\

        \cmark & \cmark & \cmark & 93.3\% & \textbf{0.530} & \textbf{14.9} & 86.1\% & 0.685 \\

    \bottomrule
  \end{tabular}
  \vspace{-1mm}
  \label{tab:abla} 
\end{table}

\paragraph{Effect of Point Selection on Enhancing Robustness}
Without point selection, noisy points directly enter the prompt generator and degrade prompt quality. Our ablation shows that prompt generation already improves robustness over full fine-tuning, and combining it with point selection brings further gains. This verifies that selecting minimally influential points effectively shields the PEFT branch from corrupted inputs.

\paragraph{Robust and Efficient Task-Oriented Prompt Generation.}
Our prompt generation module adapts the frozen pre-trained model by producing task-conditioned prompt tokens from selected points and injecting them into transformer layers before global pooling. Because the prompt branch starts from filtered points rather than raw corrupted inputs, it operates on a cleaner representation with reduced noise propagation. This enables task adaptation while preserving the robustness of the frozen backbone.

\section{Conclusion}
In this paper, we present PSFT, a point-selection fine-tuning framework for robust adaptation of 3D pre-trained models. PSFT combines minimally influential point selection, layer-wise prompt tuning on a frozen backbone, and a lightweight feature filter with stochastic residual blending. Experiments show consistent robustness gains on ModelNet-C and ModelNet40-C, together with improvements for ULIP-2 and Uni3D-B on ScanObjectNN-C, while keeping trainable parameters low.
In future work, we will evaluate PSFT under more complex and realistic 3D corruption settings to better characterize robustness and generalization in practical scenarios.

\subsubsection*{Acknowledgements}
This work was supported by the Shenzhen Science and Technology Program under
Grant KJZD20240903104400001 and was financially supported by the Research Task
Assignment Project from the Guangdong Laboratory of Artificial Intelligence and
Digital Economy (SZ), under Grant No. GML-26420004.

\bibliographystyle{splncs04}
\bibliography{references}
\clearpage
\begingroup
\setlength{\textfloatsep}{8pt plus 2pt minus 2pt}
\setlength{\floatsep}{6pt plus 2pt minus 2pt}
\setlength{\intextsep}{6pt plus 2pt minus 2pt}
\setlength{\abovecaptionskip}{4pt}
\setlength{\belowcaptionskip}{2pt}
\setlength{\aboverulesep}{0.25ex}
\setlength{\belowrulesep}{0.40ex}
\setcounter{section}{0}
\setcounter{subsection}{0}
\setcounter{figure}{0}
\setcounter{table}{0}
\setcounter{equation}{0}
\renewcommand{\thesection}{S\arabic{section}}
\renewcommand{\thesubsection}{S\arabic{section}.\arabic{subsection}}
\renewcommand{\thefigure}{S\arabic{figure}}
\renewcommand{\thetable}{S\arabic{table}}
\renewcommand{\theequation}{S\arabic{equation}}

\maketitlesupplementary

\section{Analysis of Augmentation Results}
\label{sec:supp-augmentation}
Tables~\ref{tab:suppl-modelnet-c}, \ref{tab:suppl-modelnet40-c}, and \ref{tab:suppl-scanobjectnn-c} show dataset-dependent behavior. On ModelNet-C~\cite{modelnet-c} and ModelNet40-C~\cite{modelnet40-c}, PSFT with WOLFMix~\cite{modelnet-c} achieves state-of-the-art corruption robustness among all compared settings. With Uni3D-B~\cite{uni3d}, it reaches the best mCE of 0.465 on ModelNet-C and the best mER of 12.6 on ModelNet40-C. These results indicate that augmentation and lightweight robust adaptation are highly complementary on widely used 3D corruption benchmarks~\cite{modelnet-c,modelnet40-c,adaptpoints}, while maintaining competitive clean OA.

On ScanObjectNN-C~\cite{adaptpoints}, WOLFMix is inconsistent and can hurt some backbones. Unlike synthetic benchmarks, ScanObjectNN contains real scans with heavier clutter, occlusion, and partial observations~\cite{Scanobjectnn}, making its corruption benchmark harder to optimize~\cite{adaptpoints}. Combining strong augmentation with lightweight fine-tuning may therefore destabilize convergence and weaken robustness.

\begin{table}[H]
  \setlength{\belowcaptionskip}{2pt}
  \centering
  \caption{\textbf{ModelNet-C results with augmentation.} OA is overall accuracy; mCE is mean corruption error; all other columns report CE ($\downarrow$). W.M. denotes WOLFMix~\cite{modelnet-c}; G/L indicate global/local corruptions. Other settings are from~\cite{levi2023epic,adaptpointpp,fat}; bold marks column-best values.}
  \label{tab:suppl-modelnet-c}
  \scriptsize
  \setlength{\tabcolsep}{0.7pt}
  \renewcommand{\arraystretch}{0.94}

  \begin{tabular}{@{}ll A E *{7}{E}@{}}
    \toprule
    \multirow{2}{*}{Backbone} & \multirow{2}{*}{Setting}
      & \multicolumn{1}{c}{\multirow{2}{*}{OA (\%) $\uparrow$}}
      & \multicolumn{1}{c}{\multirow{2}{*}{mCE $\downarrow$}}
      & \multicolumn{7}{c}{Corruption CE ($\downarrow$)} \\
    \cmidrule(lr){5-11}
      & & & & {Scale} & {Jitter} & {Drop-G} & {Drop-L}
      & {Add-G} & {Add-L} & {Rotate} \\
    \midrule
    \multirow{3}{*}{GDANet~\cite{xu2021gdanet}}
      & Baseline & 93.4 & 0.892 & 0.830 & 0.839 & 0.794 & 0.894
        & 0.871 & 1.036 & 0.981 \\
      & W.M. & {\bfseries 93.4} & 0.571 & 0.904 & 0.883 & 0.532 & 0.551
        & 0.305 & 0.415 & 0.409 \\
      & FAT + W.M.
        & 93.0 & 0.537 & 1.138 & {\bfseries 0.418} & 0.460 & 0.527
        & 0.281 & 0.404 & 0.530 \\
    \cmidrule(lr){1-11}
    \multirow{3}{*}{RPC~\cite{ren2022pointcloud-c}}
      & Baseline & 93.0 & 0.863 & 0.840 & 0.892 & 0.492 & 0.797
        & 0.929 & 1.011 & 1.079 \\
      & AdaptPoint++ & {--} & 0.555 & 0.957 & 0.759 & 0.367 & 0.705
        & 0.258 & {\bfseries 0.273} & 0.563 \\
      & EPiC + W.M.
        & 92.7 & 0.501 & 0.915 & 0.680 & 0.315 & {\bfseries 0.420}
        & 0.251 & 0.382 & 0.544 \\
    \cmidrule(lr){1-11}
    \multirow{3}{*}{Point-BERT~\cite{pointbert}}
      & Baseline & 92.2 & 1.098 & 0.872 & 1.503 & 0.512 & 0.903
        & 1.085 & 1.495 & 1.316 \\
      & PSFT (ours) & 91.4 & 0.790 & 1.149 & 0.503 & 0.383 & 0.961
        & 0.319 & 0.625 & 1.591 \\
      & PSFT + W.M. & 90.4 & 0.617 & 1.245 & 0.440 & 0.415 & 0.696
        & 0.329 & 0.425 & 0.772 \\
    \cmidrule(lr){1-11}
    \multirow{3}{*}{Point-MAE~\cite{pointmae}}
      & Baseline & 92.6 & 0.983 & 0.840 & 1.256 & 0.464 & 0.763
        & 1.075 & 1.265 & 1.219 \\
      & PSFT (ours) & 91.9 & 0.704 & 1.032 & 0.427 & 0.431 & 0.860
        & 0.298 & 0.502 & 1.377 \\
      & PSFT + W.M. & 89.8 & 0.630 & 1.255 & 0.475 & 0.480 & 0.696
        & 0.366 & 0.404 & 0.735 \\
    \cmidrule(lr){1-11}
    \multirow{3}{*}{ULIP-2~\cite{ulip2}}
      & Baseline & 92.1 & 0.863 & 0.851 & 1.778 & 0.492 & 0.734
        & 0.386 & 0.862 & 0.935 \\
      & PSFT (ours) & 93.3 & 0.530 & 0.819 & 0.585 & 0.298 & 0.816
        & {\bfseries 0.244} & 0.567 & 0.381 \\
      & PSFT + W.M. & 93.2 & 0.473 & 0.851 & 0.665 & {\bfseries 0.294} & 0.517
        & 0.251 & 0.396 & {\bfseries 0.340} \\
    \cmidrule(lr){1-11}
    \multirow{3}{*}{Uni3D-B~\cite{uni3d}}
      & Baseline & 92.9 & 0.746 & {\bfseries 0.777} & 1.228 & 0.415 & 0.787
        & 0.444 & 0.833 & 0.740 \\
      & PSFT (ours) & 93.4 & 0.566 & 0.862 & 0.459 & 0.327 & 0.986
        & 0.254 & 0.575 & 0.498 \\
      & PSFT + W.M. & 93.0 & {\bfseries 0.465} & 0.894 & 0.506 & 0.327 & 0.575
        & 0.254 & 0.356 & 0.344 \\
    \bottomrule
  \end{tabular}

\end{table}

\clearpage
\begin{table}[H]
  \setlength{\belowcaptionskip}{2pt}
  \centering
  \caption{\textbf{ModelNet40-C error rates with augmentation.} mER is mean error rate; W.M. denotes WOLFMix~\cite{modelnet-c}. Dens., Up., Bg., Rot., and iRBF abbreviate density, upsampling, background, rotation, and inverse RBF. Corruptions are grouped by type within one table; lower is better, and bold marks column-best values.}
  \label{tab:suppl-modelnet40-c}
  \scriptsize
  \setlength{\tabcolsep}{2.2pt}
  \renewcommand{\arraystretch}{0.88}

  \begin{tabular*}{0.90\linewidth}{@{\extracolsep{\fill}}ll R *{5}{R}@{}}
    \toprule
    \multicolumn{8}{c}{\bfseries Density corruptions} \\
    \cmidrule(lr){1-8}
    Backbone & Setting & {mER $\downarrow$} & {Occlusion} & {LiDAR}
      & {Dens.+} & {Dens.-} & {Cutout} \\
    \midrule
    PE~\cite{pointpeft} & Baseline & 22.1 & {--} & {--} & {--} & {--} & {--} \\
    \cmidrule(lr){1-8}
    OmniVec2~\cite{omnivec2} & Baseline & 14.2 & {--} & {--} & {--} & {--} & {--} \\
    \cmidrule(lr){1-8}
    \multirow{3}{*}{Point-BERT~\cite{pointbert}}
      & Baseline & 26.5 & 54.7 & 73.8 & 10.8 & 13.4 & 13.6 \\
      & PSFT (ours) & 20.0 & 56.9 & 70.9 & 9.2 & 10.6 & 9.6 \\
      & PSFT + W.M. & 16.9 & 56.3 & 60.0 & 10.2 & 10.1 & 9.8 \\
    \cmidrule(lr){1-8}
    \multirow{3}{*}{Point-MAE~\cite{pointmae}}
      & Baseline & 23.9 & 53.5 & 74.8 & 10.7 & 11.3 & 11.9 \\
      & PSFT (ours) & 18.3 & 55.5 & 66.6 & 9.0 & 10.7 & 10.1 \\
      & PSFT + W.M. & 16.9 & 55.3 & 58.4 & 10.5 & 10.7 & 10.4 \\
    \cmidrule(lr){1-8}
    \multirow{3}{*}{ULIP-2~\cite{ulip2}}
      & Baseline & 24.1 & 51.4 & 80.0 & 9.7 & 10.8 & 10.9 \\
      & PSFT (ours) & 14.9 & 44.6 & 53.3 & 7.5 & 8.8 & 9.0 \\
      & PSFT + W.M. & 13.9 & 40.8 & 51.9 & {\bfseries 7.0} & {\bfseries 7.6} & {\bfseries 7.5} \\
    \cmidrule(lr){1-8}
    \multirow{3}{*}{Uni3D-B~\cite{uni3d}}
      & Baseline & 18.9 & 44.2 & 58.2 & 9.6 & 10.3 & 11.3 \\
      & PSFT (ours) & 14.2 & 41.2 & {\bfseries 42.4} & 8.1 & 9.8 & 10.5 \\
      & PSFT + W.M. & {\bfseries 12.6} & {\bfseries 40.1} & 42.6 & 7.5 & 7.7 & 7.7 \\
    \midrule
    \multicolumn{8}{c}{\bfseries Noise corruptions} \\
    \cmidrule(lr){1-8}
    Backbone & Setting & {mER $\downarrow$} & {Uniform} & {Gaussian}
      & {Impulse} & {Up.} & {Bg.} \\
    \midrule
    PE~\cite{pointpeft} & Baseline & 22.1 & 14.1 & 17.1 & 15.7 & 13.7 & 17.9 \\
    \cmidrule(lr){1-8}
    OmniVec2~\cite{omnivec2} & Baseline & 14.2 & {--} & {--} & {--} & {--} & {--} \\
    \cmidrule(lr){1-8}
    \multirow{3}{*}{Point-BERT~\cite{pointbert}}
      & Baseline & 26.5 & 17.5 & 22.8 & 46.7 & 23.8 & 36.7 \\
      & PSFT (ours) & 20.0 & 11.1 & 11.2 & 10.4 & 10.2 & 9.9 \\
      & PSFT + W.M. & 16.9 & 11.3 & 11.2 & 9.8 & 10.7 & 10.5 \\
    \cmidrule(lr){1-8}
    \multirow{3}{*}{Point-MAE~\cite{pointmae}}
      & Baseline & 23.9 & 15.7 & 18.5 & 33.0 & 18.8 & 35.5 \\
      & PSFT (ours) & 18.3 & 10.5 & {\bfseries 10.3} & 9.4 & 9.6 & 9.6 \\
      & PSFT + W.M. & 16.9 & 11.9 & 11.6 & 9.7 & 10.7 & 11.2 \\
    \cmidrule(lr){1-8}
    \multirow{3}{*}{ULIP-2~\cite{ulip2}}
      & Baseline & 24.1 & 29.9 & 35.7 & 18.7 & 31.9 & 12.9 \\
      & PSFT (ours) & 14.9 & 12.0 & 12.8 & 8.2 & 9.5 & 7.6 \\
      & PSFT + W.M. & 13.9 & 14.9 & 15.8 & {\bfseries 7.7} & 10.4 & {\bfseries 7.5} \\
    \cmidrule(lr){1-8}
    \multirow{3}{*}{Uni3D-B~\cite{uni3d}}
      & Baseline & 18.9 & 15.1 & 18.9 & 20.2 & 21.6 & 14.1 \\
      & PSFT (ours) & 14.2 & {\bfseries 10.0} & 10.8 & 8.9 & 9.7 & 7.6 \\
      & PSFT + W.M. & {\bfseries 12.6} & 10.8 & 11.1 & 7.8 & {\bfseries 9.5} & 8.0 \\
    \midrule
    \multicolumn{8}{c}{\bfseries Transformation corruptions} \\
    \cmidrule(lr){1-8}
    Backbone & Setting & {mER $\downarrow$} & {Rotation} & {Shear}
      & {FFD} & {RBF} & {iRBF} \\
    \midrule
    PE~\cite{pointpeft} & Baseline & 22.1 & {--} & {--} & {--} & {--} & {--} \\
    \cmidrule(lr){1-8}
    OmniVec2~\cite{omnivec2} & Baseline & 14.2 & {--} & {--} & {--} & {--} & {--} \\
    \cmidrule(lr){1-8}
    \multirow{3}{*}{Point-BERT~\cite{pointbert}}
      & Baseline & 26.5 & 25.3 & 13.3 & 14.5 & 15.9 & 14.9 \\
      & PSFT (ours) & 20.0 & 30.7 & 16.9 & 15.5 & 14.1 & 13.5 \\
      & PSFT + W.M. & 16.9 & 12.6 & 10.2 & 10.3 & 10.0 & 10.0 \\
    \cmidrule(lr){1-8}
    \multirow{3}{*}{Point-MAE~\cite{pointmae}}
      & Baseline & 23.9 & 22.7 & 11.5 & 13.1 & 13.9 & 13.0 \\
      & PSFT (ours) & 18.3 & 24.3 & 12.9 & 12.9 & 12.0 & 11.3 \\
      & PSFT + W.M. & 16.9 & 11.7 & 10.5 & 10.2 & 10.3 & 10.0 \\
    \cmidrule(lr){1-8}
    \multirow{3}{*}{ULIP-2~\cite{ulip2}}
      & Baseline & 24.1 & 18.6 & 10.5 & 12.1 & 14.7 & 13.8 \\
      & PSFT (ours) & 14.9 & 7.6 & 7.1 & 10.4 & 12.5 & 11.9 \\
      & PSFT + W.M. & 13.9 & 7.2 & {\bfseries 6.8} & 7.3 & 7.7 & 7.8 \\
    \cmidrule(lr){1-8}
    \multirow{3}{*}{Uni3D-B~\cite{uni3d}}
      & Baseline & 18.9 & 13.4 & 8.8 & 11.7 & 13.4 & 12.6 \\
      & PSFT (ours) & 14.2 & 9.0 & 7.5 & 11.1 & 13.3 & 12.7 \\
      & PSFT + W.M. & {\bfseries 12.6} & {\bfseries 7.0} & 7.1
        & {\bfseries 7.3} & {\bfseries 7.7} & {\bfseries 7.7} \\
    \bottomrule
  \end{tabular*}
\end{table}

\clearpage
\begin{table}[H]
  \centering
  \caption{\textbf{ScanObjectNN-C results with augmentation.} OA and mCE denote overall accuracy and mean corruption error. W.M. denotes WOLFMix~\cite{modelnet-c}; PointAugment~\cite{pointaugment} and PointMixup~\cite{pointmixup} are additional augmentations. G and L denote global and local corruptions. Bold denotes the best result in each column.}
  \label{tab:suppl-scanobjectnn-c}
  \scriptsize
  \setlength{\tabcolsep}{0.8pt}
  \renewcommand{\arraystretch}{0.94}

  \begin{tabular}{@{}ll A E *{7}{E}@{}}
    \toprule
    \multirow{2}{*}{Backbone} & \multirow{2}{*}{Setting}
      & \multicolumn{1}{c}{\multirow{2}{*}{OA (\%) $\uparrow$}}
      & \multicolumn{1}{c}{\multirow{2}{*}{mCE $\downarrow$}}
      & \multicolumn{7}{c}{Corruption CE ($\downarrow$)} \\
    \cmidrule(lr){5-11}
      & & & & {Scale} & {Jitter} & {Drop-G} & {Drop-L}
      & {Add-G} & {Add-L} & {Rotate} \\
    \midrule
    \multirow{4}{*}{PointNeXt~\cite{pointnext}}
      & Baseline & 87.3 & 0.921 & 0.803 & 1.079 & 0.807 & 0.942
        & 0.944 & 0.875 & 0.995 \\
      & W.M. & {\bfseries 87.7} & 0.869 & 0.819 & 1.193 & 0.897 & 0.780
        & 0.893 & {\bfseries 0.800} & 0.700 \\
      & PointAugment & 87.4 & 0.894 & 0.808 & 1.000 & 0.775 & 0.858
        & 0.976 & 0.850 & 0.993 \\
      & PointMixup & 88.2 & 0.904 & 0.801 & 1.035 & 0.799 & 0.927
        & 0.876 & 0.837 & 1.056 \\
    \cmidrule(lr){1-11}
    \multirow{3}{*}{Point-BERT~\cite{pointbert}}
      & Baseline & 83.4 & 0.931 & 0.618 & 1.088 & 0.587 & 0.736
        & 0.661 & 1.630 & 1.195 \\
      & PSFT (ours) & 79.4 & 0.970 & 0.898 & 0.954 & 0.608 & 0.937
        & 0.493 & 1.612 & 1.288 \\
      & PSFT + W.M. & 65.6 & 1.235 & 0.711 & 1.217 & 0.931 & 1.327
        & 0.793 & 2.075 & 1.592 \\
    \cmidrule(lr){1-11}
    \multirow{3}{*}{Point-MAE~\cite{pointmae}}
      & Baseline & 84.3 & 0.895 & 0.609 & 0.952 & 0.569 & 0.726
        & 0.550 & 1.722 & 1.135 \\
      & PSFT (ours) & 77.6 & 1.014 & 0.960 & 0.974 & 0.701 & 0.990
        & 0.580 & 1.568 & 1.326 \\
      & PSFT + W.M. & 64.3 & 1.333 & 0.727 & 1.344 & 0.997 & 1.389
        & 0.876 & 2.291 & 1.704 \\
    \cmidrule(lr){1-11}
    \multirow{3}{*}{ULIP-2~\cite{ulip2}}
      & Baseline & 86.2 & 0.777 & 0.538 & 1.096 & 0.524 & {\bfseries 0.637}
        & 0.376 & 1.348 & 0.918 \\
      & PSFT (ours) & 86.1 & {\bfseries 0.685} & 0.621 & 0.960 & {\bfseries 0.431} & 0.670
        & {\bfseries 0.333} & 1.079 & {\bfseries 0.700} \\
      & PSFT + W.M. & 82.3 & 0.786 & {\bfseries 0.412} & 1.189 & 0.585 & 0.842
        & 0.424 & 1.278 & 0.775 \\
    \cmidrule(lr){1-11}
    \multirow{3}{*}{Uni3D-B~\cite{uni3d}}
      & Baseline & 85.8 & 0.801 & 0.509 & 1.085 & 0.532 & 0.680
        & 0.365 & 1.502 & 0.933 \\
      & PSFT (ours) & 86.4 & 0.702 & 0.626 & {\bfseries 0.871} & 0.458 & 0.733
        & 0.341 & 1.079 & 0.805 \\
      & PSFT + W.M. & 80.5 & 0.833 & 0.448 & 1.094 & 0.622 & 0.947
        & 0.435 & 1.427 & 0.861 \\
    \bottomrule
  \end{tabular}
\end{table}

\par\noindent
\begin{minipage}{\linewidth}
These results quantify the dataset dependence. Averaged over four backbones, WOLFMix lowers ModelNet-C mCE by 0.101 and ModelNet40-C mER by 1.8 points, but raises ScanObjectNN-C mCE by 0.204 and reduces clean OA by 9.2 points. All backbones follow this negative ScanObjectNN-C trend, establishing an augmentation--dataset mismatch whose cause--clutter, occlusion, partial observations, or interaction with scan noise--requires controlled study.
\end{minipage}\par

\endgroup
\end{document}